\documentclass[sigconf]{acmart}
\AtBeginDocument{%
  \providecommand\BibTeX{{%
    \normalfont B\kern-0.5em{\scshape i\kern-0.25em b}\kern-0.8em\TeX}}}

\usepackage{balance}

\setcopyright{acmcopyright}

\copyrightyear{2022}
\acmYear{2022}
\setcopyright{acmcopyright}\acmConference[MM '22]{Proceedings of the 30th ACM International Conference on Multimedia}{October 10--14, 2022}{Lisboa, Portugal} \acmBooktitle{Proceedings of the 30th ACM International Conference on Multimedia (MM '22), October 10--14, 2022, Lisboa, Portugal}
\acmPrice{15.00}
\acmDOI{10.1145/3503161.3547851}
\acmISBN{978-1-4503-9203-7/22/10}

\begin{document}

\title{Prompting for Multi-Modal Tracking}

\author{Jinyu Yang}

\affiliation{%
  \institution{Southern University of Science and Technology}
  \institution{University of Birmingham}
   \country{}
}
\email{jinyu.yang96@outlook.com}

\author{Zhe Li}
\affiliation{%
  \institution{Southern University of Science and Technology}
   \country{}
}
\email{liz8@mail.sustech.edu.cn}

\author{Feng Zheng} \authornote{Corresponding author.}
\affiliation{%
  \institution{CSE \& RITAS, Southern University of Science and Technology}
   \country{}
}
\email{f.zheng@ieee.org}

  \country{}

\author{Ale\v{s} Leonardis}
\affiliation{%
 \institution{University of Birmingham}
  \country{}
}
\email{a.leonardis@cs.bham.ac.uk}

\author{Jingkuan Song}
\affiliation{%
  \institution{University of Electronic Science and Technology of China}
   \country{}
}
\email{jingkuan.song@gmail.com}




\renewcommand{\shortauthors}{Jinyu Yang et al.}

\begin{abstract}
Multi-modal tracking gains attention due to its ability to be more accurate and robust in complex scenarios compared to traditional RGB-based tracking.
Its key lies in how to fuse multi-modal data and reduce the gap between modalities.
However, multi-modal tracking still severely suffers from data deficiency, thus resulting in the insufficient learning of fusion modules.
Instead of building such a fusion module, in this paper, we provide a new perspective on multi-modal tracking by attaching importance to the multi-modal visual prompts.
We design a novel multi-modal prompt tracker (ProTrack), which can transfer the multi-modal inputs to a single modality by the prompt paradigm.
By best employing the tracking ability of pre-trained RGB trackers learning at scale, our ProTrack can achieve high-performance multi-modal tracking by only altering the inputs, even without any extra training on multi-modal data.
Extensive experiments on 5 benchmark datasets demonstrate the effectiveness of the proposed ProTrack.
\end{abstract}





\maketitle

\section{Introduction}
Object tracking is to localize an arbitrary object in a video sequence given the object description in the first frame. 
Recent years witness a great spur of object tracking research, especially in color mode.
Large-scale datasets \cite{fan2019lasot,Huang_2019,muller2018trackingnet}, worldwide challenges~\cite{kristan2019seventh,Kristan2020a,kristan2021ninth}, and flourishing novel algorithms \cite{keeptrack,transt,dimp,prdimp} boost this area.
At the same time, multi-modal tracking, \textit{e.g.}, Visible+Depth (RGB-D), Visible+Thermal (RGB-T), and Visible+Event (RGB-E), opens up opportunities for more accurate and robust tracking in complex scenarios.
For example, depth cameras provide spatial information to distinguish the object from the background, thermal cameras have a strong penetration ability to particulate matters like smog and fog, and event cameras are free from motion blurs 
and low illumination.
Therefore, in complex scenarios, especially when trackers fail in color views, multi-modal information can strongly complementary.

However, unlike the success in color-based object tracking, we note that there exists a significant development gap in multi-modal tracking.
High-performing data-driven models are lacking for multi-modal tracking, due to the deficiency of large-scale training datasets.
For example, a comparison among RGB tracking datasets and multi-modal tracking ones is shown in Table~\ref{tab1}, clearly showing that the multi-modal tracking datasets are orders of magnitude smaller than the RGB counterparts.
Also, the resolution of multi-modal data is relatively smaller than the RGB ones.
In addition, multi-modal datasets also suffer from low-quality problems like desynchronization and misalignment between different modalities or sensors.


\begin{table}
\begin{center}
\caption{Dataset comparison between RGB tracking datasets and multi-modal ones. ``M'' denotes million. ``Resolution'' indicates the maximum resolution.}\label{tab1}
\setlength\tabcolsep{3pt}
\begin{tabular}{|l|c|c|c|c|c|c|}
\hline
Dataset &Year&Modality & Videos & Frames &Resolution\\ 
\hline
LaSOT\cite{fan2019lasot}&2019 &RGB & 1,400 &3.52M &$1280\times720$ \\
GOT-10k\cite{8922619} &2019 &RGB & 10,000&1.5M &$1920\times1080$ \\
TrackingNet\cite{muller2018trackingnet} &2018 & RGB & 31,000&14M & $1280\times720$ \\
\hline
DepthTrack\cite{yan2021depthtrack} &2021 &RGB+D &200 &294,600 &$640\times480$\\
VisEvent\cite{wang2021viseventbenchmark} &2021&RGB+E& 820 &371,127 &$346\times260$ \\
LasHeR\cite{2021LasHeR} &2021&RGB+T &1,224&734,800 & $960\times576$\\
\hline
\end{tabular}
\end{center}
\end{table}

To handle the data-hungry bottleneck, compared to manually collecting the multi-modal data for deep model designing, a more efficient way is to best exploit the state-of-the-art pre-trained RGB tracking models.
Current state-of-the-arts in multi-modal tracking already have many attempts on it mostly with the ``pre-trained RGB baselines + multi-modal data fine-tuned'' paradigm.
In RGB-D tracking, trackers use pre-trained RGB trackers as strong baselines and show promising performance \cite{yan2021depthtrack}.
In the VisEvent benchmark, Wang et al. \cite{wang2021viseventbenchmark} proposed a series of baselines by embedding event information into RGB trackers and re-trained on its training set.
However, these models can only learn a small amount of new data with modality gaps, resulting in incomplete learning of the cross-modal fusion mechanism.
Further, inevitably, such a modality gap will confuse models that were originally well-learned on large-scale RGB datasets.
Even though we can have large-scale cross-modal datasets in the future, current research clearly illustrates that over-training on a new dataset also potentially causes a knowledge forgetting problem. 
Therefore, a question is raised: how can we effectively utilize both the large-scale RGB knowledge and the complementary information from non-color modalities?

\begin{figure}
  \includegraphics[width=0.9\linewidth]{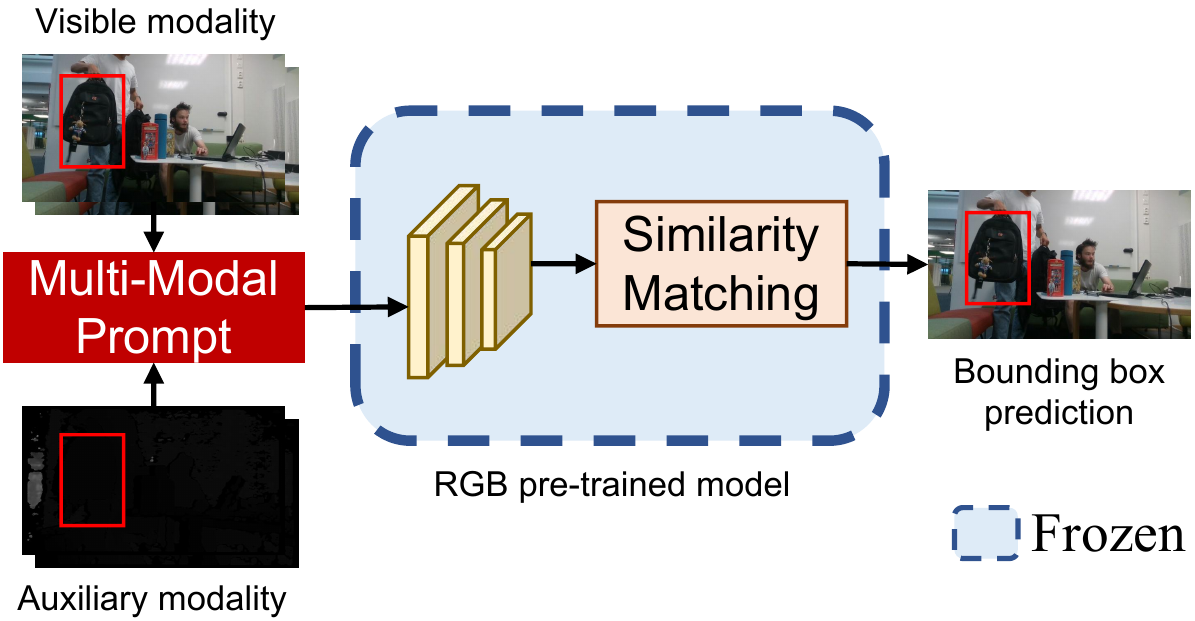}
  \caption{How our multi-modal prompt works. Given a frozen, pre-trained RGB tracker, we expect it can perform well on multi-modal tracking tasks with only a modality-agnostic prompt on the test videos.}
  \label{fig:test}
\end{figure}

\begin{figure*}
  \includegraphics[width=\linewidth]{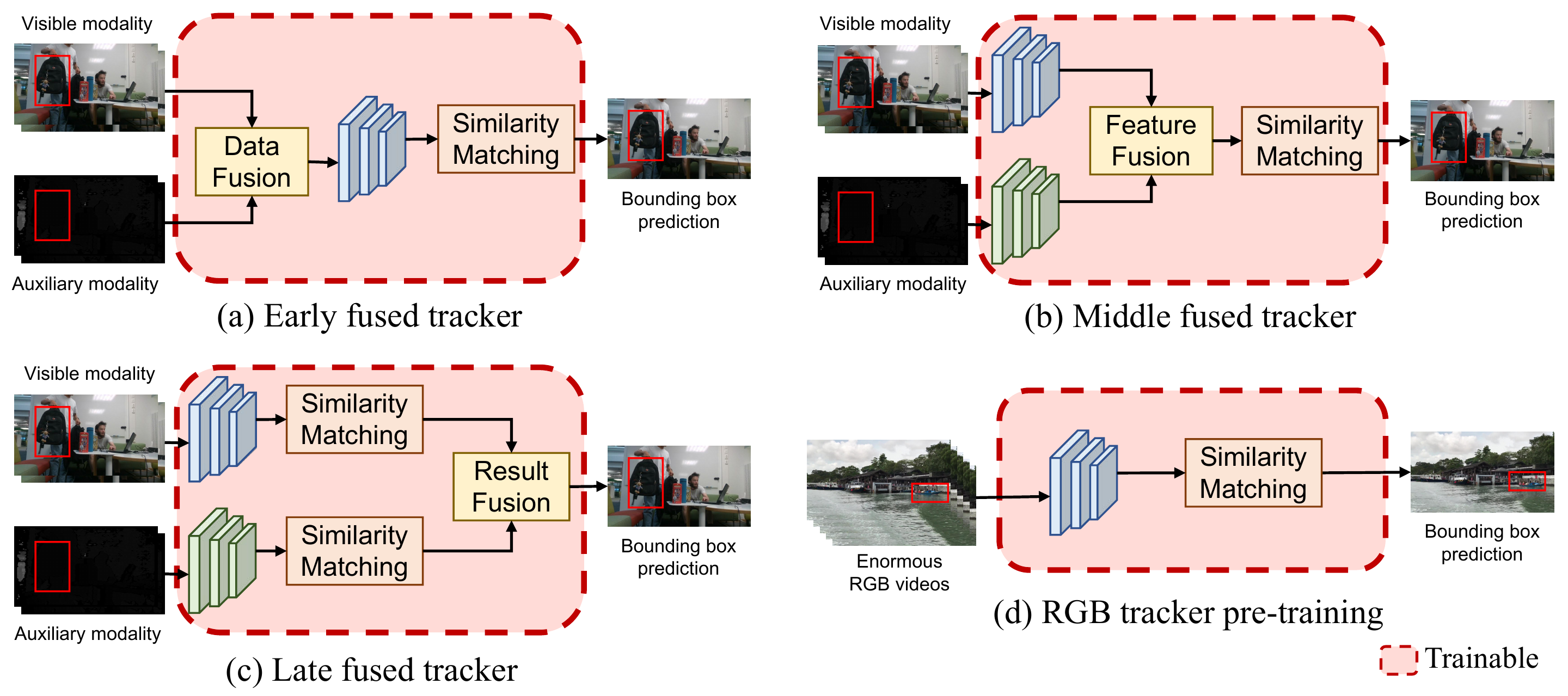}
  \caption{Comparison between our proposed method (d) and the existing ones (a, b, and c) during the training process.}
  \label{fig:train}
\end{figure*}

To achieve high-performing multi-modal tracking more efficiently, we explore a different route in this paper.
Firstly, we observe that there are visible and auxiliary modalities in multi-modal tracking.
Tracking performance highly depends on the models' discriminative ability in color views, which can be gained from large-scale data training, while auxiliary modalities are effective in specific scenarios.
Moreover, due to the data sources/sensors, the auxiliary modality is less informative.
Inappropriately importing auxiliary modalities may receive more kicks than halfpence.
Drawing inspiration from the recent advances in prompting in Natural Language Processing (NLP), we therefore ask, 
can we still use auxiliary modalities but without the inappropriate fine-tuning process?
To this end, we propose a new simple but efficient prompt method for multi-modal object tracking tasks, namely multi-modal Prompt Tracker (ProTrack).
Instead of altering or fine-tuning the pre-trained model itself, we directly modify multi-modal inputs to adapt the state-of-the-art RGB trackers as shown in Fig.~\ref{fig:test}.
The comparison between our ProTrack and the existing three types of fusion strategies for multi-modal tracking methods is shown in Fig.~\ref{fig:train}.
Regardless of the stage at which classical methods fuse multiple data, there are two main differences compared to our methods.
On one hand, these models have to be fine-tuned on new datasets for new tasks, while our model can be merely trained once on RGB videos.
On the other hand, since these models are fine-tuned for only one type of data, they cannot be used in different multi-modal scenarios, while our one model can handle multiple tasks, simultaneously.

By selecting the appropriate prompts, we can manipulate the model behavior so that the pre-trained model itself can be used to predict the desired output, without any additional task-specific or modality-specific training.
Unlike efforts on vision-and-language tasks with textual prompts, we introduce the tracking applications of this promising paradigm with multi-modal visual prompts.
By transferring the multi-modal data to colored ones, the invisible auxiliary modality information can be leveraged by RGB trackers as color-based referential markers.
We adopt the prompt engineering in 5 multi-modal object tracking benchmarks.
Enormous experiments verify its effectiveness. 
Our contributions are three-fold:
\begin{itemize}
\item {We present a novel prompt learning paradigm for multi-modal tracking, in which both the large-scale RGB knowledge and the complementary information are effectively utilized.
To the best of our knowledge, this is the first attempt on prompt learning in multi-modal tracking areas.
}
\item {We present a principled solution to cross-modal prompt configurations for various kinds of multi-modal tracking but without the inappropriate fine-tuning process.
}
\item {We unify different multi-modal object tracking tasks into a prompting framework and conduct comprehensive experiments on different scenarios which demonstrate the effectiveness of ProTrack.
}
\end{itemize}

\section{Related Work}

\subsection{Multi-Modal Tracking}
Tracking through joint information from different modalities is an intuitive idea, whose effectiveness has been verified by many existing works.
Preliminary explorations focused on auxiliary modality representation to assist the trackers on specific occasions.
For example, early methods use depth and thermal maps for occlusion handling \cite{liu2018context} and scale estimation \cite{2018Real}.
Similarly, event features are extracted for target re-detection after model drifting \cite{event1}.
With the development of deep learning, data-driven models appeared, which devote to fusing different modalities with deep features.
Wang \textit{et al.} \cite{wang2021viseventbenchmark} construct a series of baseline trackers by fusing visible frames and event flows on RGB tracking baselines and re-train them on the proposed training dataset.
Li \textit{et al.} \cite{Chenglong2018Fusing} concatenate deep features of visible and thermal frames and adaptively fuse them with the proposed FusionNet to get robust feature representations.
Qian \textit{et al.} \cite{2019DAL} propose embedded depth-aware convolution in RGB tracking baselines to obtain robust target estimation.
At the same time, a rising problem in those areas is the deficiency of large-scale training datasets.
To deal with the data-hungry bottleneck, besides collecting new datasets, Zhang \textit{et al.} \cite{zhang2019multi} generate a large-scale synthetic RGB-T dataset from 8,335 videos, and Yan \textit{et al.} \cite{yan2021depthtrack} use synthetic RGB-D data generated from LaSOT~\cite{fan2019lasot} and COCO~\cite{lin2014microsoft}.

\subsection{Prompt Learning}
``Pre-train, prompt, and predict'' paradigm is replacing the ``pre-train, fine-tune'' one in NLP area.
In the prompt paradigm, downstream tasks are reformulated to look more like the solved ones with the help of textual prompts \cite{prompt}.
Instead of adapting pre-trained language models to downstream tasks via fine-tuning, prompting learning can reach high performance even in the few-shot or zero-shot settings.
Extending from NLP, Pre-trained vision-language models also show promising capabilities in many tasks. %
CLIP \cite{clip} is trained from scratch on a dataset of 400 million (image, text) pairs collected from the internet, and evaluated on 30 different existing computer vision datasets.
Despite CLIP, multiple works in various research fields have emerged.
For example, Zhou \textit{et al.} \cite{coop} proposed context optimization (CoOp) which learns continuous soft prompts that perform well for downstream tasks. 
CLIP-Adapter \cite{clipadapter} is then proposed to conduct fine-tuning with feature adapters on both visual and language branches.

\subsection{Prompting for Vision Tasks}
There have been preliminary attempts on prompt with images in vision tasks.
CPT \cite{cpt} converts visual grounding into a fill-in-the-blank problem by creating visual prompts with color-based co-referential markers in both images and text.
Very recently, Jia \textit{et al.} \cite{vpt} proposed VPT which applied visual prompts to vision backbones on 24 classification tasks.
Bahng \textit{et al.} \cite{vp} proposed visual prompting which learns a task-specific image perturbation to adapt the pre-trained models to downstream tasks.
With only changing a few pixels, the visual prompting shows surprising effectiveness.
To the best of our knowledge, there are no prompt paradigms designed for semantic-agnostic multi-modal vision tasks.

\section{Method}

\subsection{Prompt Formulation}

In the classical paradigm of NLP, in general, models are first pre-trained on large-scale datasets and then fine-tuned on downstream data to adapt to new tasks.
The data and supervision are often different between the two stages, thus leading to forgetting or under-fitting problems.
Instead of updating the model, a new prompt paradigm is proposed, using text templates to modify the data to fit the input of the pre-trained model.
The reason to fix these models is that they have learned at scale, seeing more concepts that cannot be provided by downstream datasets.
The key component of prompting is how to design the templates (textual prompts), which can reduce the distribution gap between the pre-trained data and downstream data.

\subsection{Intuition}

\begin{table}
\centering
\caption{Terminology and notation of our proposed multi-modal prompting methods.}\label{termin}
\begin{tabular}{|c|c|p{5cm}|}
\hline
Name & Notation & Description\\
\hline
Input&$X =\{V,A\}$ & Multi-modal frames.\\
\hline
Prompt & $f(X)$ & A function that converts the input into \\
function&  & a specific form.\\
\hline
Prompt & $X' = f(X)$ & Prompted RGB frames generated by prompt function.\\
\hline
Model & $tracker\{X'\}$ & A pre-trained model for object tracking with input $X'$.\\
\hline
Output&$B$& Predicted target bounding box.\\
\hline
\end{tabular}
\end{table}

Similarly, the key to our multi-modal prompts is to reformulate multi-modal tracking into a single-modal tracking problem.
To this end, our ProTrack establishes fine-grained connections between single-modal videos and multi-modal videos. 
Table~\ref{termin} shows how we define the prompting method in multi-modal tracking.
The multi-modal input $X=\{V,A\}$ is converted to $X'\in \mathcal{R}^{3\times H \times W}$ after the prompt function $f(X)$, where $H$ and $W$ are the height and width of the input image, respectively. 
Here we denote visible modality as $V$ and auxiliary modality as $A$.
In ProTrack, the pre-trained model is from the RGB tracking area while the downstream tasks are multi-modal object tracking tasks.
Specifically, the ProTrack framework consists of two components: \\
1) a multi-modal prompt that transfers the multi-modal video sequences into visible single-modal ones; \\ %
2) a pre-trained model $tracker\{X\}$ that has a strong discriminative ability in the visible tracking area.

We then illuminate how we generate multi-modal prompts and reformulate the tracking problems.

\subsection{Multi-Modal Prompt Design} 

Given that the prompt specifies the task, choosing a proper prompt has a large effect on not only the accuracy but also which task the model performs in the first place.

\textbf{Pre-trained tracker.}
In this work, given a multi-modal video in multi-modal tracking, we usually have the Visible Modality (RGB) and Auxiliary Modality (\textit{e.g.}, depth, event, or thermal).
The original multi-modal tracking process can be formulated as:
\begin{equation}
     tracker: \{X_t, X_1, B_1\}~\rightarrow~{B_t}
\end{equation}
where $B_t$ denotes the predicted bounding box and $B_1$ is the bounding box supervision in the first frame. 
Thus, as shown in Fig. \ref{fig:train} (a,b,c), the model $tracker$ will be trained using multi-modal data $X=\{V,A\}$. 

Instead, as shown in Fig. \ref{fig:train}(d), our model $tracker$ is trained merely using existing large-scale RGB data to associate the objects in different frames. 
Thus, the input size of the tracker is fixed to $3\times H \times W$.
Therefore, no fusion module is required, neither in the training stage nor in the testing stage. 
We consider a variant of the spatial-temporal transformer \cite{yan2021learning} as the baseline tracker.
For more details, please refer to Sec. \ref{pretrainmodel}.
The intuitive consideration is that RGB videos are more readily available than collecting multi-model datasets, so we have already had more RGB videos.
The immediate benefit of using an RGB tracker is that the tracker has seen many challenges in learning from big data.
If we can transfer other challenges to these already-seen challenges by introducing new modality data, the association performance will be improved further. 

\textbf{Prompt.}
Obviously, if we want to use the pre-trained models without any updates, we need to transfer the multi-modal data $X=(V,A)$ into the solution $3 \times H \times W$.
With only modifying the trackers' input, our multi-modal prompt function can be formulated as:
\begin{equation}\label{prompt}
     f(V_t, A_t) = \lambda * Color(A_t) + (1 - \lambda) * Color(V_t),
\end{equation}
where $\lambda$ is a parameter. 
$Color(*)$ denotes dyeing function of different modalities. 
If the data $V$ has one channel, the result $Color(V)$ will have three channels. 
But if the data $V$ is the RGB image, the operation $Color(V)$ does nothing.
Equipped with multi-modal prompts, it is then straightforward to apply pre-trained RGB trackers on prompted videos, without any extra training:
\begin{equation}
     tracker: \{f(V_t, A_t), f(V_1, A_1), B_1\}~\rightarrow~{B_t}.
\end{equation}
It is worth noting that the pre-trained RGB tracker has not seen any multi-modal data.


\subsection{Why Multi-Modal Prompt Works?}
Compared with a special design to read and understand different channels, our solution is more straightforward. 
By dyeing salient colors or highlighting the auxiliary modality on brightness, the prompting in our ProTrack can reach state-of-the-art performance. In summary, the following three aspects clearly explain why the prompting strategy is very effective for such types of tasks.

\textbf{Color image is more informative.}
Conventional cameras can capture informative images and videos.
Multi-modal information is proposed for vision tasks at first due to their sensitivity to specific activities.
At the same time, the information provided by the sensors is more focused and less informative.
In existing works, researchers expect the visible modality and auxiliary modality from two sensors can cooperate with each other to achieve more reliable object tracking.
However, inappropriate fusion or combination between different modalities may damage the information volume, which finally has side effects on tracking performance.
The key to exploiting multi-modal information is preserving the visible information at most and embedding the useful parts of the auxiliary information at the same time.
Thus, the prompt paradigm then provides us a new perspective to adapt the tasks to the pre-trained models, in which our design will not damage the discriminative ability of the original frameworks.
In other words, to make multi-modal prompt work, we have to guarantee the parameter $\lambda$ is relatively very small.

\textbf{Prompting reduces the gap between distributions.}
\begin{figure}
  \includegraphics[width=0.9\linewidth]{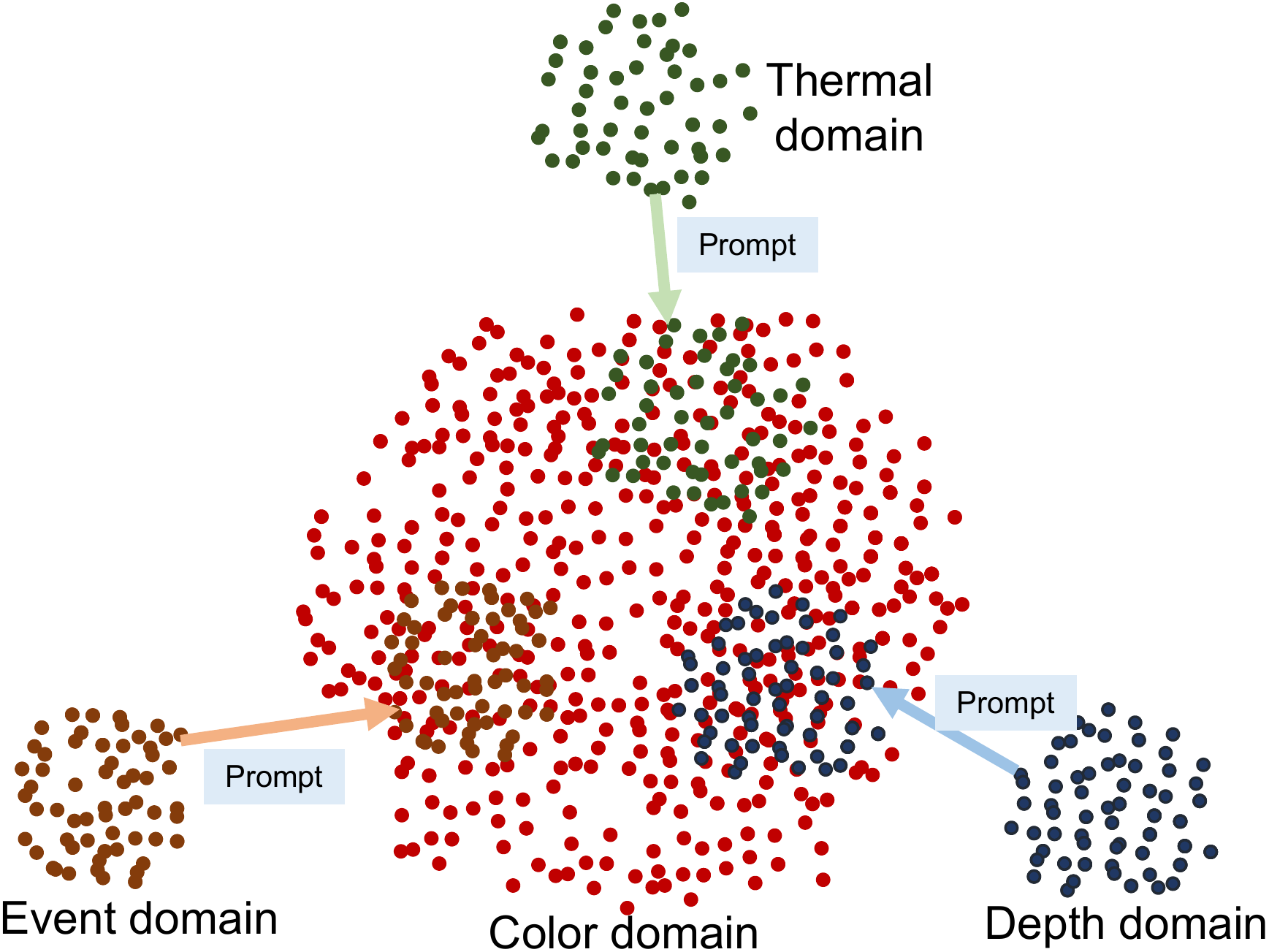}
  \caption{Multi-modal prompts in the data domain reduce the distribution gap between the pre-trained data (red) and downstream data.}
  \label{fig:distribution}
\end{figure}
Obviously, the data distribution of auxiliary modality is very different from that of the counterpart RGB videos.
Thus, naturally, feeding this data directly into a model that learns only from RGB videos fails to achieve good performance.
As shown in Fig. \ref{fig:distribution}, the purpose of prompting is to transfer the data field of auxiliary modality to the main field of RGB images. 
The newly generated samples capture the inherent characteristics of both modalities, and since they lie within the distribution region learned by the model, the model can make better predictions.
The essential benefit is that the distribution gap between the different modalities has narrowed considerably.

\textbf{Large-scale data plays a vital role.}
The original purpose of learning new fused data is to enhance the ability to meet new challenges.
However, due to limited data volumes, not only this goal is not achieved, but the fusion module is not fully learned.
Furthermore, fine-tuning somewhat destroys the associating ability learned from RGB videos.
That's why the main problem we have is that the current fusion models cannot work well or even match the performance of using only RGB trackers.
In contrast, our prompting strategy can successfully avoid such a paradox.
On the one hand, auxiliary data is used, and thus new challenges can be solved.
On the other hand, trackers pre-trained on large-scale data remain unchanged, thus avoiding the knowledge forgetting problem.
Thus, we believe our prompting is a very good compromise under the condition that currently we have no large-scale multi-modal data.
Nevertheless, in the future, we also believe that collecting more multi-modal data can better address this issue.

\section{Experiments}
We evaluate the proposed ProTrack for a wide range of downstream multi-modal tracking tasks with pre-trained RGB trackers. 
We first describe our experimental settings in Sec.~\ref{setup}, including the downstream tasks and parameter settings.
Then a brief introduction to the pre-trained model is given in Sec.~\ref{pretrainmodel}.
Then we demonstrate the effectiveness of our method on the downstream multi-modal tracking tasks in Sec.~\ref{result}. 
We also systematically study how different design choices would affect performance in Sec.~\ref{ablation}, which leads to an improved understanding of our approach.

\subsection{Experimental Settings} \label{setup}
To verify the effectiveness of our proposed ProTrack, we select the following downstream tracking tasks:
%

\begin{itemize}
\item \textbf{RGB-Depth object tracking.} We compare trackers on the popular CDTB \cite{Lukezic_2019_ICCV} and DepthTrack \cite{yan2021depthtrack}.
\item \textbf{RGB-Thermal object tracking.} We compare trackers on the large-scale LasHeR \cite{2021LasHeR} and RGBT234 \cite{rgbt234}.
\item \textbf{RGB-Event object tracking.} We compare trackers on the largest VisEvent \cite{wang2021viseventbenchmark}.
\end{itemize}

In experiments, we empirically set $\lambda = 0.05$.
For color choices, we use JET colormaps for depth maps and thermal images by default, while for event data, we simply use the event images transformed from event flows. 
All experiments are run on a single 32GB Tesla V100 GPU.

\begin{figure}
  \includegraphics[width=\linewidth]{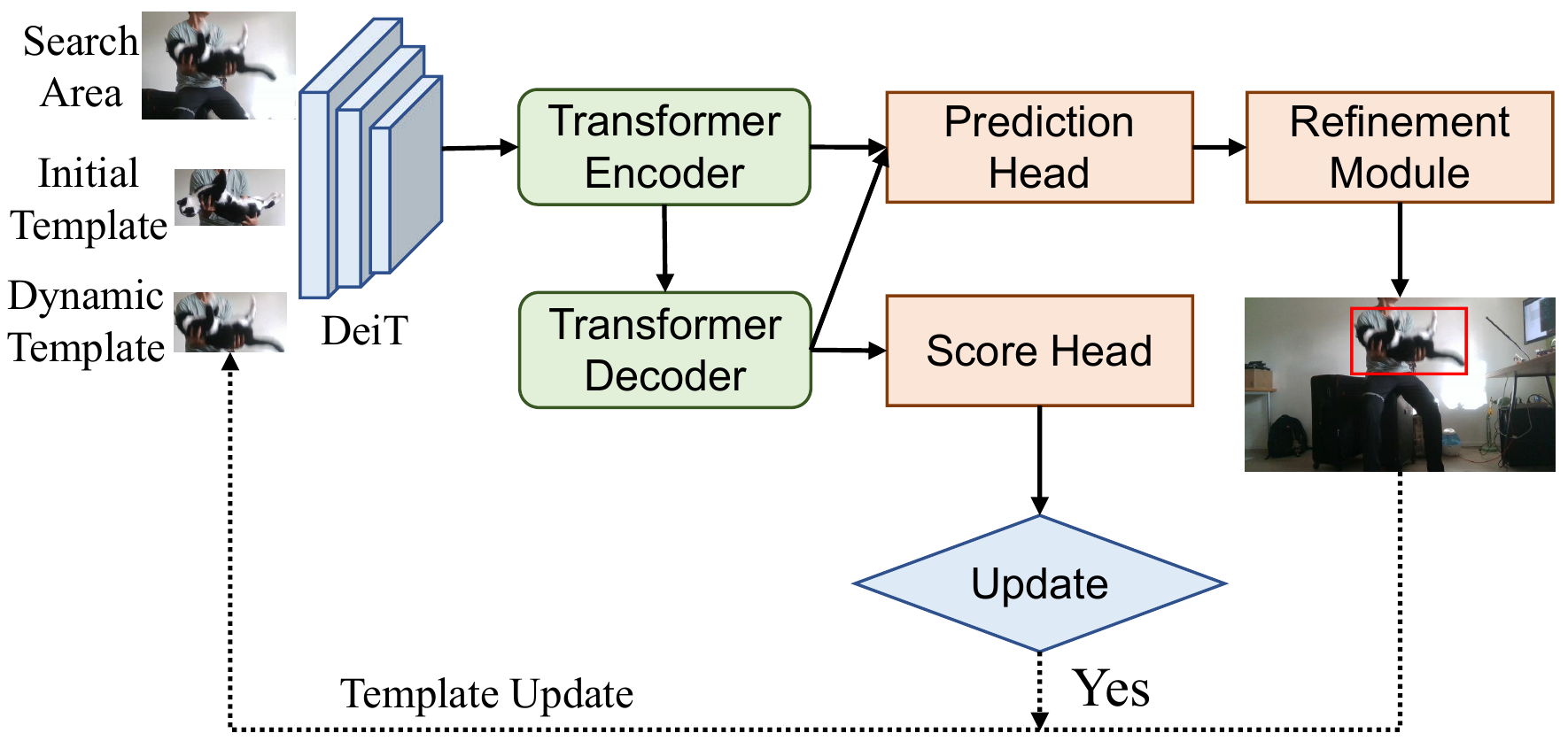}
  \caption{Architecture of the pre-trained model STARK.}
  \label{fig:stark}
\end{figure}

\subsection{Pre-trained Model Selection} \label{pretrainmodel}
Since we aim to fully employ the discriminative ability of RGB pre-trained models, the choice of pre-trained models is of vital importance.
Here we choose a variant of spatial-temporal transformer \cite{yan2021learning}, which leads the leaderboard of RGB object tracking benchmarks.
In this paper, we specifically use ``STARK'' denoting the variant model we selected.

Here we briefly introduce the pre-trained model we selected.
STARK is based on spatial-temporal transformer architecture.
As shown in Fig.~\ref{fig:stark}, STARK contains the following main components: backbone, encoder, decoder, prediction head, and score head.
For feature extraction, STARK utilizes DeiT \cite{DeiT} to strengthen the deep features.
The encoder-decoder is based on DETR \cite{detr}.
With inputting both the initial template and dynamic template, the encoder extracts the spatial-temporal features by modeling the correlation in both spatial and temporal dimensions.
The decoder takes a single target query to predict a bounding box.
The prediction head firstly takes the search region features from the encoder's output, then computes the similarity between the search region features and the output embedding from the decoder.
The similarity is used to enhance the search area features and then estimate the probability distribution of the box corners.
With a refinement module based on AlphaRefine \cite{yan2020alpha}, we finally get the bounding box prediction.
The output is also used as a dynamic template to enhance the target template in most cases.
Since there are mostly long-term settings in multi-modal tracking tasks, which means the targets may disappear and reappear during the tracking process, STARK utilizes a score head to determine whether the dynamic template should be updated.

Specifically, STARK is trained in two stages.
In the first stage, the whole network, except for the score head, is trained end-to-end.
In the second stage, only the score head is optimized with BCE loss.
The training data consists of the training sets from the aforementioned LaSOT~\cite{fan2019lasot}, GOT-10K \cite{Huang_2019}, COCO2017 \cite{lin2014microsoft}, and TrackingNet \cite{muller2018trackingnet}.
Top performance on multiple tracking benchmarks demonstrates the discriminative ability of this model.
For more details, please refer to \cite{yan2021learning}.

\subsection{Main Results}\label{result}

\begin{table*}[!t]
\caption{Overall performance on the CDTB dataset \cite{Lukezic_2019_ICCV}.}
\centering
\label{tbl_cdtb}
\begin{tabular}{l|ccccccccc}
\hline
 Method  & DS-KCF\cite{camplani2015real} &CA3DMS\cite{liu2018context} &CSR\_RGBD++\cite{csr} &OTR\cite{otr} &DAL\cite{2019DAL} &TSDM\cite{zhao2021tsdm} &DeT\cite{yan2021depthtrack} &STARK\cite{yan2021learning} & \textbf{ProTrack}\\
\hline
Pr& 0.036  &0.271 &0.187 &0.336 & 0.662 & 0.578& 0.674 &0.740 &\textbf{0.747}\\
Re &0.039 &0.284 &0.201 &0.364 & 0.565& 0.541&0.642 &0.765 &\textbf{0.767} \\
F-score &0.038 &0.259 & 0.194&0.312 &0.592 &0.559 &0.657 &0.752 &\textbf{0.757}\\
\hline
\end{tabular}
\end{table*}

\begin{table*}[!t]
\caption{Overall performance on DepthTrack test set \cite{yan2021depthtrack}.}
\centering
\label{tbl_depthtrack}
\begin{tabular}{l|cccccccc}
\hline
 Method  & DS-KCF-shape\cite{2016DS} &CA3DMS\cite{liu2018context} &CSR\_RGBD++\cite{csr} &DAL\cite{2019DAL} &TSDM\cite{zhao2021tsdm} &DeT\cite{yan2021depthtrack} &STARK\cite{yan2021learning} & \textbf{ProTrack}\\
\hline
Pr & 0.023 & 0.212 & 0.113  &0.478 & 0.393& 0.560 &0.558 &\textbf{0.583}  \\
Re &0.023  &0.216 & 0.115  &0.390 & 0.376 & 0.506 &0.543 &\textbf{0.573}\\
F-score  &0.023  & 0.214 & 0.114   &0.421 &0.384 &0.532 & 0.550 &\textbf{0.578}\\
\hline
\end{tabular}
\end{table*}

\textbf{CDTB dataset} \cite{Lukezic_2019_ICCV} consists of 80 long-term RGB-D video sequences for evaluation, covering a wide range of challenges in RGB-D tracking.
In the CDTB dataset, target objects disappear and reappear frequently.
Tracking Precision ($Pr$) and Recall ($Re$) are computed under a series of confidence thresholds.
F-score is obtained by $F =\frac{2Re \times Pr}{Re + Pr}$.
As shown in Table~\ref{tbl_cdtb}, we compare our ProTrack with existing stare-of-the-art RGB-D trackers on it.
As reported, with the same ResNet-50 backbones, our ProTrack outperforms DeT \cite{yan2021depthtrack} by 10\%.
Besides, ProTrack reaches a new state-of-the-art F-score of 75.7\%, which also surpasses the STARK without prompting.

\textbf{DepthTrack dataset} \cite{yan2021depthtrack} is a large-scale long-term RGB-D tracking benchmark.
We evaluate RGBD trackers on the DepthTrack test set, which contains 50 long RGB-D video sequences.
DepthTrack dataset uses the same protocols as the CDTB dataset.
Table~\ref{tbl_depthtrack} presents that our ProTrack surpasses all previous state-of-the-art trackers, obtaining a new state-of-the-art F-score of 57.8\%.
Also, without multi-modal prompts, the F-score of STARK is 0.550, while our ProTrack has an improvement of 2.8\% thanks to the informative prompts.

\begin{figure}
  \includegraphics[width=\linewidth]{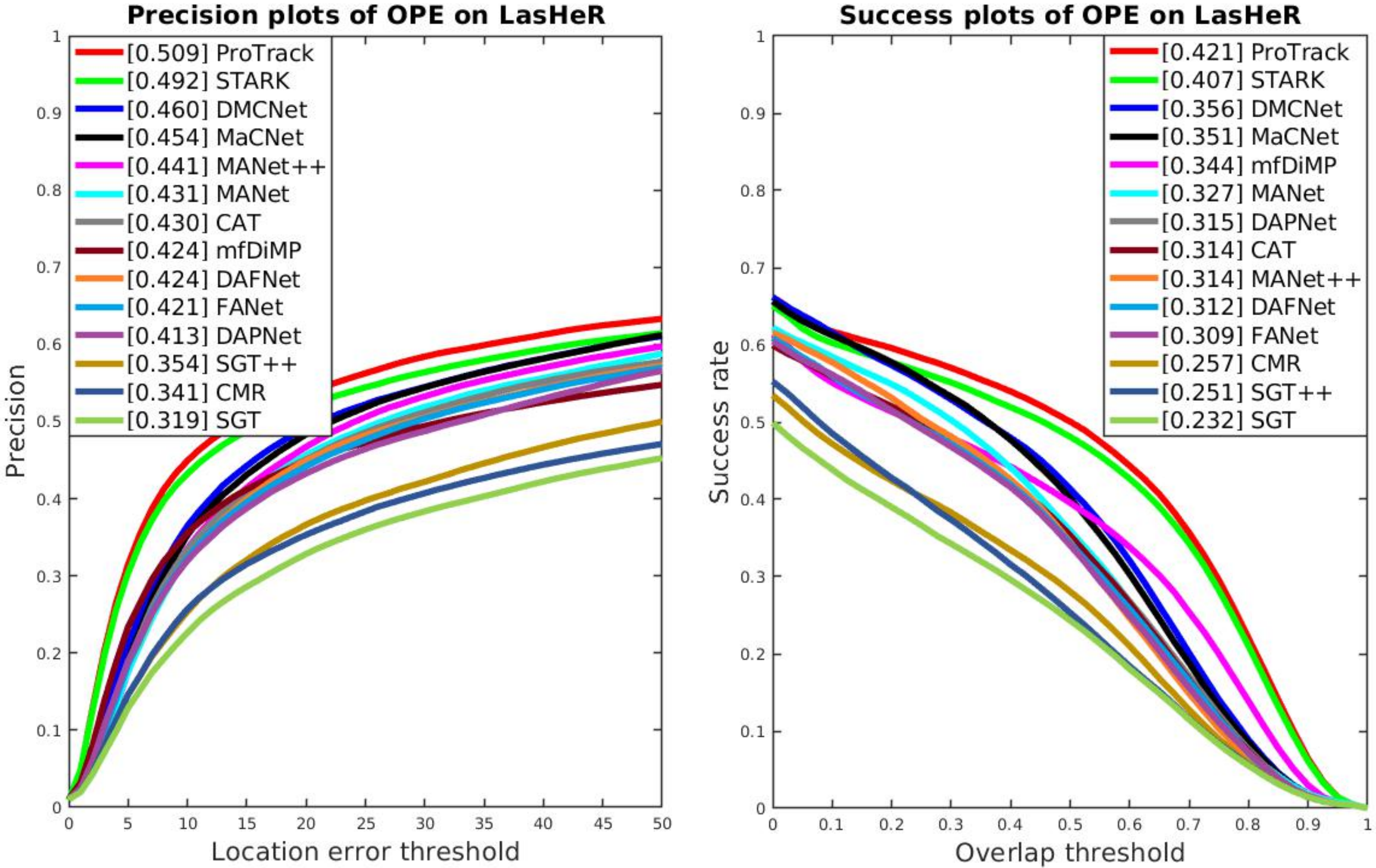}
  \caption{Overall performance on the LasHeR test set \cite{2021LasHeR}.}
  \label{fig:lasher}
\end{figure}

\textbf{LasHeR dataset} \cite{2021LasHeR} is a large-scale high-diversity benchmark for short-term RGB-T tracking.
LasHeR equips the standard tracking performance metrics including Precision Plot and Success Plot. 
We evaluate trackers on the testing subset which contains 245 video sequences.
Here we compare our ProTrack with existing stare-of-the-art RGB-T trackers.
Note that the compared models are trained on the LasHeR training set.
The results are reported in Fig.~\ref{fig:lasher}.
As shown, our ProTrack shows the top performance of 50.9/41.9 (precision/success), outperforming the well-performing DMCNet \cite{lu2022duality} by 4.9\%/6.3\%.

\begin{figure}
  \includegraphics[width=\linewidth]{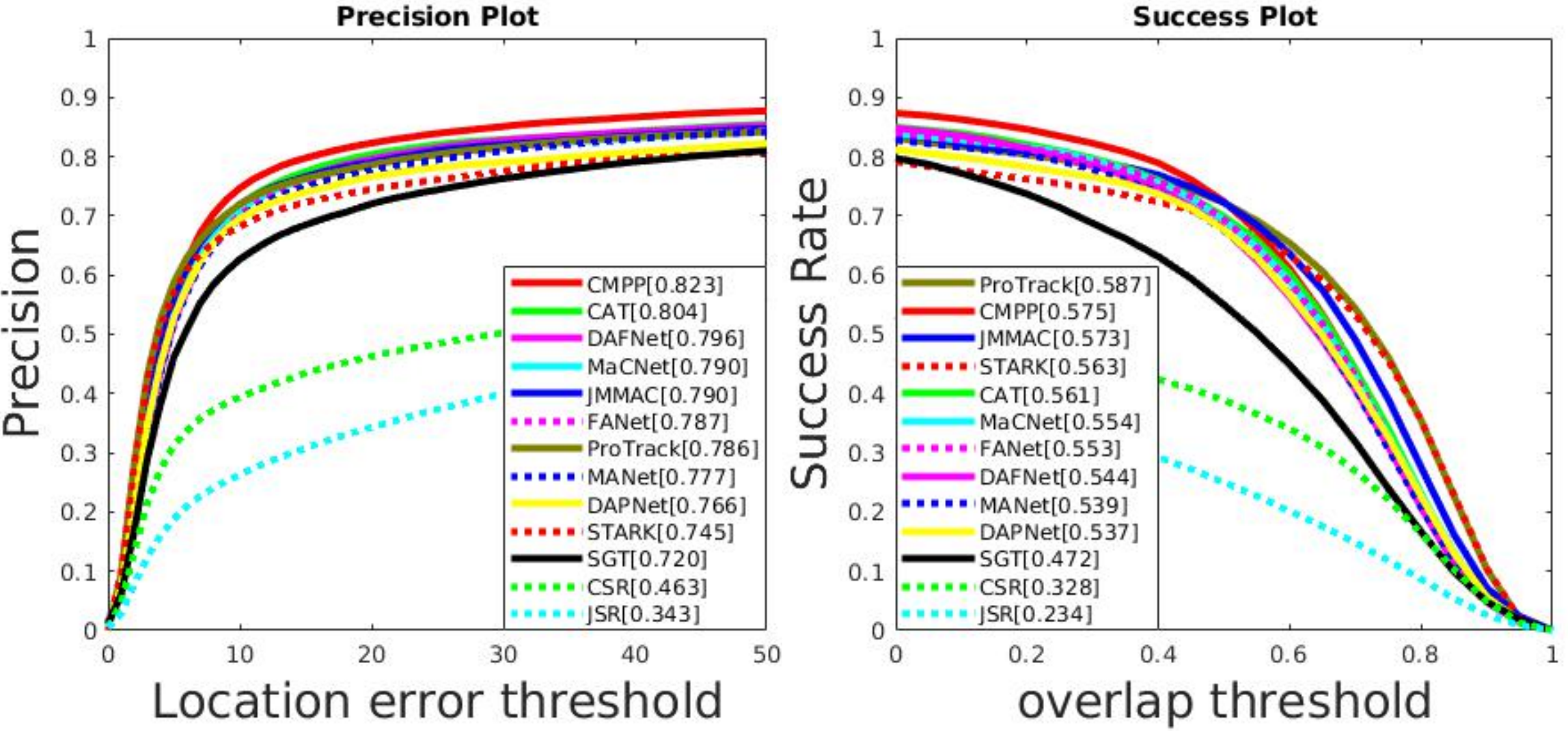}
  \caption{Overall performance on the RGBT234 dataset \cite{rgbt234}.}
  \label{fig:rgbt234}
\end{figure}

\textbf{RGBT234 dataset} \cite{rgbt234} is a large-scale RGB-T tracking dataset for performance evaluation, which contains 234 videos and 116.6k image pairs.
Here we compare ProTrack with state-of-the-art RGB-T trackers.
Comparison results are shown in Fig.~\ref{fig:rgbt234}.
Without pre-training or adaptive learning on RGB-T data, our ProTrack can obtain a competitive precision rate of 78.6\%.
Moreover, ProTrack reaches the top success rate of 58.7\%, which beats the well-designed RGB-T trackers.
Compared to the STARK which is used for comparison, our ProTrack has improvements of 4.1\% and 2.4\% on precision and success rate, respectively.
Till now, we observe that our ProTrack shows better on the success plots compared to the precision ones.
Note that the precision plot requires manually setting thresholds for location error, which might be too tolerant for trackers to obtain higher precision.
While the success plot reflects the real overlap ratio, which can be fairer than the precision one.

\begin{figure}
  \includegraphics[width=\linewidth]{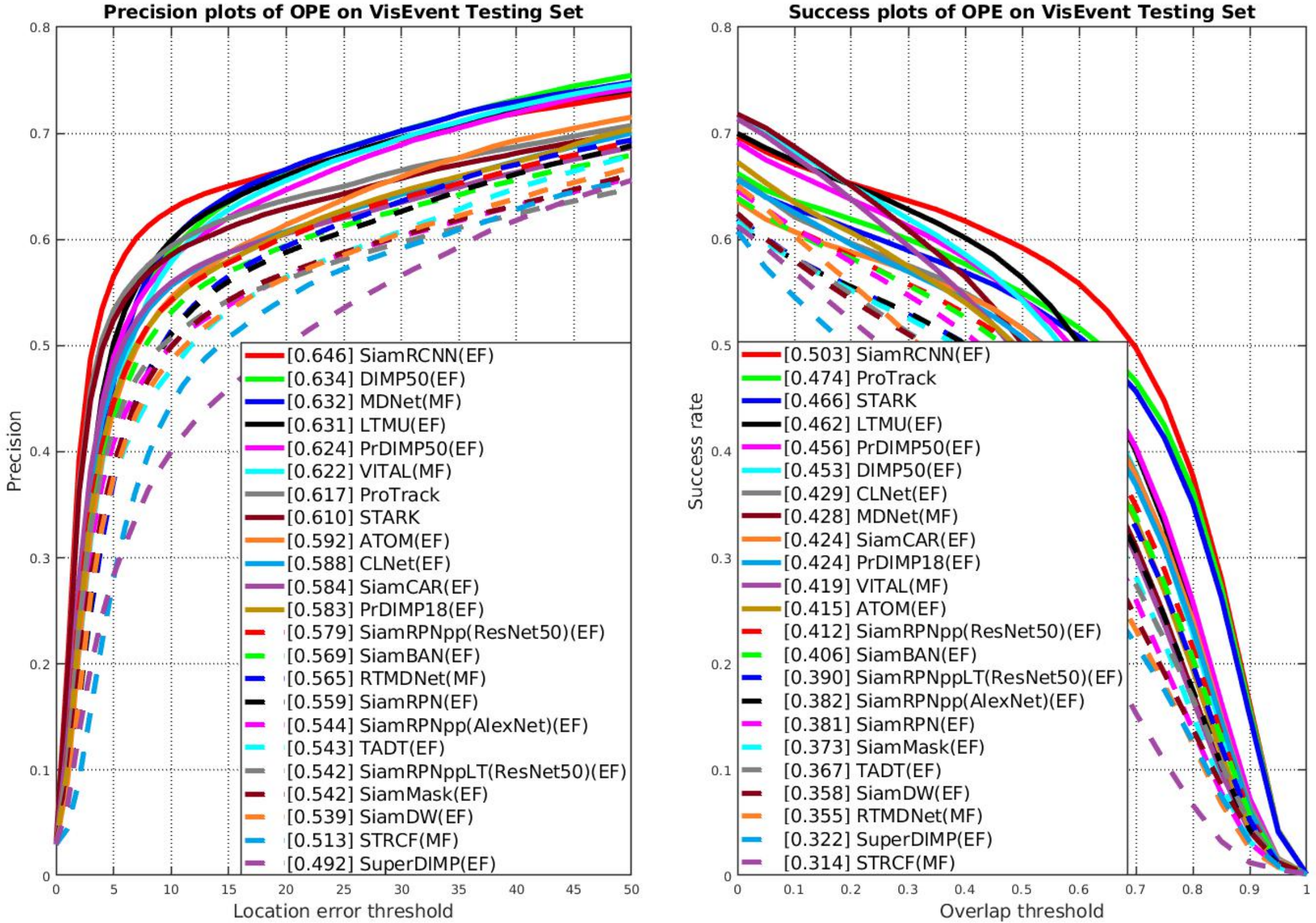}
  \caption{Overall performance on VisEvent test set \cite{wang2021viseventbenchmark}.}
  \label{fig:visevent}
\end{figure}


\textbf{VisEvent dataset} \cite{wang2021viseventbenchmark} is a large-scale Visible-Event benchmark.
We evaluate trackers on the VisEvent test subset (320 videos).
Two standard metrics are adopted for the evaluation of tracking performance, including Precision Plot and Success Plot. 
We involve the multiple baseline methods \cite{wang2021viseventbenchmark} to compare, which keep state-of-the-art on VisEvent.
Note that the compared trackers are dedicated to visible-event object tracking and fine-tuned on the VisEvent training subset.
Our ProTrack still performs on par with current SOTAs.
Except for without domain-specific training, another reason for our sub-optimal performance is that the event modality after prompting can only provide positive/negative ($+1/-1$) on some pixels, which is much less informative compared to the depth or thermal maps.

\subsection{Ablation Study}\label{ablation}
We ablate different prompt design choices on the large-scale benchmarks for ablation study.
For DepthTrack \cite{yan2021depthtrack}, we use F-score for comparison.
For LasHeR \cite{2021LasHeR} and VisEvent \cite{wang2021viseventbenchmark}, we use precision rate and success rate.

\textbf{Effect of different modalities.}
To investigate the contributions of different modalities, we evaluate the tracking performance with single-modal inputs.
Results are reported in Table~\ref{tbl_ab1}.
``Default'' denotes the input data we used with multi-modal prompts.
Auxiliary information is transferred to colormaps for evaluation.
As shown, the tracker performs better in visible modality compared to auxiliary modalities, and the performance is highly dependent on visible information.
While they are not comparable to the ones with our default multi-modal prompts, which effectively work on multi-modal data.

\textbf{Effect of color choices.}
As we choose colormaps for representation, colors choices are the key components in tracking the performance of ProTrack. 
Specifically, we compare different color choices as shown in Table~\ref{tbl_ab2}.
By default, the colormaps follow the JET style for covering more colors in RGB domain.
Since event cameras only capture the changing pixels, there are only red or blue pixels by default in RGB-Event experiments.
``RED'' is the single-color colormap covering (0,0,0) to (255,0,0).
``GRAY'' indicates that we only use the grayscale maps.
As reported, the default settings can achieve better performance compared to the single colored ones.
It indicates that more colors bring more information and are more similar to the natural RGB image color distribution.

\textbf{Effect of parameter $\lambda$.}
In practice, the hyperparameter $\lambda$, which indicates the trade-off between different modalities, is crucial in ProTrack.
To investigate the effect of $\lambda$, we evaluate ProTrack with different $\lambda$, as shown in Table.~\ref{tbl_ab3}.
When $\lambda=0$, there are no prompts and the input remains the visible modality only.
As reported, the performance of ProTrack increases first and then decreases with $\lambda$ improves on DepthTrack and VisEvent.
This can be explained by the that, a tiny $\lambda$ can preserve strong discriminative ability from visual appearances, but will undermine the visibility of the auxiliary modality, and vice versa. 
In addition, we observe that the performance on LasHeR keeps going higher with $\lambda$ going larger, indicating that the thermal maps are more informative or more color-like, and thus they can provide more complementary compared to depth and event information.
Thus, choosing a proper $\lambda$ is essential in our ProTrack.

\begin{table}[!t]
\caption{Ablation study on different modalities.}
\centering
\label{tbl_ab1}
\begin{tabular}{|l|c|cc|cc|}
\hline
  &\multicolumn{1}{c|}{DepthTrack\cite{yan2021depthtrack}} & \multicolumn{2}{c|}{LasHeR\cite{2021LasHeR}} & \multicolumn{2}{c|}{VisEvent\cite{wang2021viseventbenchmark}} \\ 
  \cline{2-6}
Modality &\multicolumn{1}{c|}{F-score} & \multicolumn{1}{c}{Pre} & \multicolumn{1}{c|}{Suc}  & \multicolumn{1}{c}{Pre} & \multicolumn{1}{c|}{Suc} \\
\hline
Default & 0.578  &0.509 &0.419& 0.617 &0.474\\
Visible & 0.550 & 0.492 &0.405 &0.610 &0.466\\
Auxiliary & 0.297 &0.349 &0.289 &0.411 &0.277\\
\hline
\end{tabular}
\end{table}

\begin{table}[!t]
\caption{Ablation study on color choices.}
\centering
\label{tbl_ab2}
\begin{tabular}{|l|c|cc|cc|}
\hline
  &\multicolumn{1}{c|}{DepthTrack\cite{yan2021depthtrack}} & \multicolumn{2}{c|}{LasHeR\cite{2021LasHeR}} & \multicolumn{2}{c|}{VisEvent\cite{wang2021viseventbenchmark}} \\ 
  \cline{2-6}
Color &\multicolumn{1}{c|}{F-score} & \multicolumn{1}{c}{Pre} & \multicolumn{1}{c|}{Suc}  & \multicolumn{1}{c}{Pre} & \multicolumn{1}{c|}{Suc} \\
\hline
Default & 0.578 &0.509 &0.419& 0.617 &0.474 \\
RED &0.564 &0.486 &0.401 &0.596 & 0.443\\
GRAY & 0.561 &0.501 &0.410 &0.546 & 0.396\\
\hline
\end{tabular}
\end{table}

\begin{table}[!t]
\caption{Ablation study on parameter $\lambda$. \textbf{Bold} denotes the highest score.}
\centering
\label{tbl_ab3}
\begin{tabular}{|l|c|cc|cc|}
\hline
  &\multicolumn{1}{c|}{DepthTrack\cite{yan2021depthtrack}} & \multicolumn{2}{c|}{LasHeR\cite{2021LasHeR}} & \multicolumn{2}{c|}{VisEvent\cite{wang2021viseventbenchmark}} \\ 
  \cline{2-6}
$\lambda$  &\multicolumn{1}{c|}{F-score} & \multicolumn{1}{c}{Pre} & \multicolumn{1}{c|}{Suc}  & \multicolumn{1}{c}{Pre} & \multicolumn{1}{c|}{Suc} \\
\hline
0& 0.550 &0.492 &0.405 &0.610 &0.466\\
0.01 & 0.557 &0.495 &0.407 &0.612 &0.469\\
0.05(Default) & \textbf{0.578} &0.509 &0.419 & \textbf{0.617} &\textbf{0.474}\\
0.1 & 0.537 &0.527 &0.436 &0.611 & 0.468\\
0.2 & 0.499 &\textbf{0.531} &\textbf{0.439} &0.606 &0.454\\
\hline
\end{tabular}
\end{table}

\begin{table*}[!t]
\caption{ProTrack with more pre-trained models. Performance on DepthTrack \cite{yan2021depthtrack} is shown according to F-score. ``\_P'' denotes tracking performance after prompting.}
\centering
\small
\label{tbl_pretrain}
\setlength\tabcolsep{3pt}
\begin{tabular}{cc| cc|c c|cc| cc }
\hline
 ATOM\cite{atom} &ATOM\_P &DiMP\cite{dimp} &DiMP\_P &PrDiMP\cite{prdimp} & PrDiMP\_P &TransT\cite{transt} & TransT\_P & KeepTrack\cite{keeptrack} & KeepTrack\_P \\
\hline
  0.313 & 0.349 &0.377  &0.431 &0.392 &0.402 & 0.489& 0.504 &0.509 & 0.542\\
\hline
\end{tabular}
\end{table*}

\section{Analysis and Discussion}

In this section, we conduct a deep analysis of ProTrack for a better understanding of its working mechanism from various perspectives.

\subsection{Visualization}
\begin{figure}
  \includegraphics[width=\linewidth]{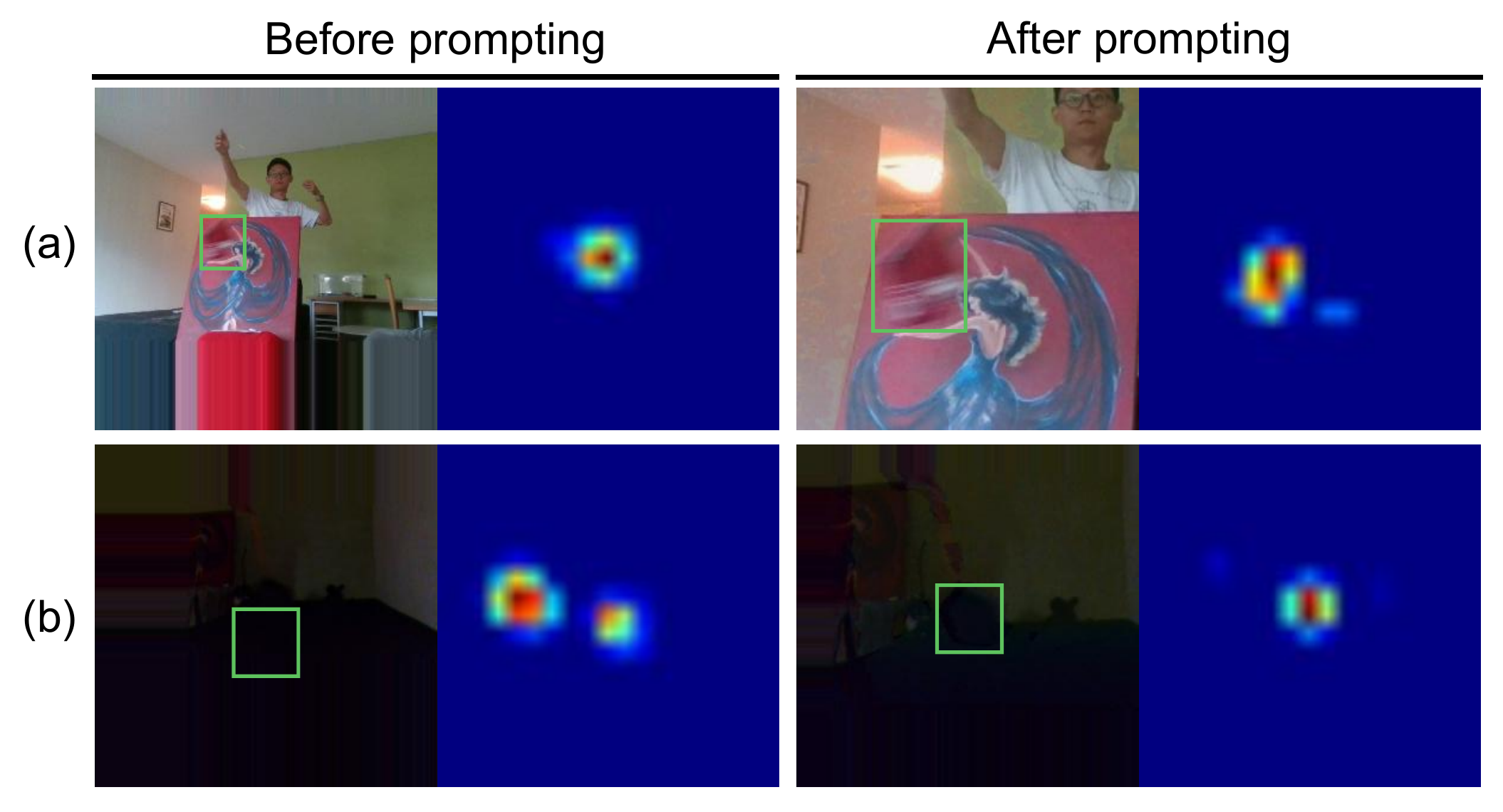}
  \caption{Visualized comparison of the score maps in search regions with/without prompting. The groundtruth bounding box is shown in green.}
  \label{fig:visualize}
\end{figure}

To investigate how the multi-modal prompts work during the tracking process, we visualize the search regions and corresponding feature maps
before and after prompting in two representative tracking frames, as shown in Fig.~\ref{fig:visualize}.
As shown, we observe that our ProTrack has much cleaner and unambiguous target localization than the one without prompting and leads to an accurate bounding box estimation.
While the one without prompts often produces multiple local maxima for distractors, our methods are able to almost fully suppress these.
An important design enabling this is the prompts with auxiliary information.
Although it is almost invisible to see the auxiliary information after prompting, the feature maps tend to distinguish the objects from background distractors due to the change in overall data distribution.
The visualized examples verify the hypothesis shown in Fig.~\ref{fig:distribution}.
Our multi-modal prompts can transfer the invisible auxiliary modality into the visible one by transferring the auxiliary data into the color domain.

\subsection{ProTrack with More Pre-trained Models}
As ProTrack aims to leverage the information learned from enormous RGB data, many alternatives can be used as pre-trained models in our ProTrack framework.
Here we choose some RGB trackers to simply verify the effectiveness of multi-modal prompts.
The representative trackers include ATOM \cite{atom}, TransT \cite{transt}, DiMP \cite{dimp}, PrDiMP \cite{prdimp}, and KeepTrack \cite{keeptrack}, which are pre-trained on large-scale RGB datasets, \textit{e.g.}, LaSOT~\cite{fan2019lasot}, GOT-10K \cite{Huang_2019}, COCO \cite{lin2014microsoft}, and TrackingNet \cite{muller2018trackingnet}.
We report the results of trackers' F-score on DepthTrack without and with prompting, respectively.
Compared results are shown in Table~\ref{tbl_pretrain}.
The RGB trackers show improved performance on multi-modal tracking after prompting.
Specifically, DiMP and KeepTrack get improvements of 5.4\% and 3.3\% after prompting, respectively.
Thus, ProTrack can continuously benefit from the strong discriminative ability of pre-trained models.

\subsection{Beyond Dual-Modal tracking}
As we claim that our ProTrack is modality-agnostic, its applications can be broader.
For example, we can apply ProTrack to triple-modal tracking by modifying Eq.~\ref{prompt} to:
\begin{equation}\label{triple}
     f(V, A_1, A_2) = \alpha * Color(A_1) + \beta * Color(A_2) + \gamma * Color(V),
\end{equation}
where $\alpha, \beta, \gamma$ are parameters, satisfying $\alpha + \beta + \gamma = 1$.
And, $V, A_1, A_2$ denote visible modality and two auxiliary modalities.
With prompting, information from different modalities can be converged to the visible modality and get assistance from RGB pre-trained tracking models.
Unfortunately, triple-modal tracking is not explored yet.
We hope our work will provide a straightforward solution for this promising direction.

\subsection{Failed Cases}

\begin{figure}
  \includegraphics[width=\linewidth]{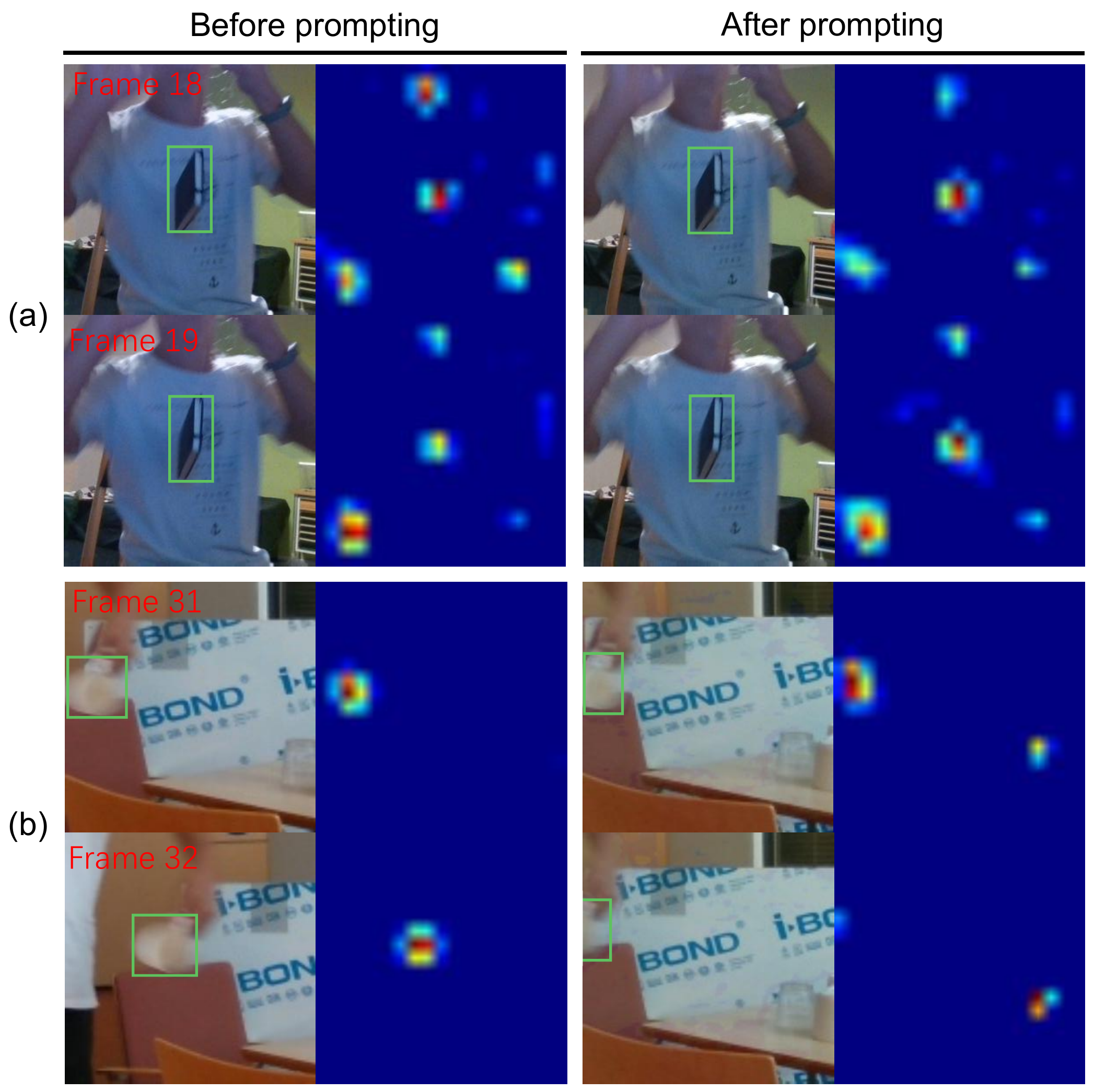}
  \caption{Visualization of failure cases of our tracker. The groundtruth bounding box is shown in green.}
  \label{fig:failed}
\end{figure}

Fig.~\ref{fig:failed} shows failed cases of our tracker. 
In particular, it shows the adjacent frames in two sequences containing the groundtruth in search regions and the corresponding score maps of the pre-trained tracker with or without prompting.
We can see that the location of highest scores is not matching the groundtruth.
Overall, our tracker typically fails due to two reasons: 
1) The tracker's discriminative ability is limited to distinguishing the object in some challenging scenarios.
2) The input image is disturbed by the multi-modal prompts.
For the former reason, complex sequences exist and challenge the trackers, as shown in Fig.~\ref{fig:failed}(a).
This can be solved by the improvements of pre-trained RGB trackers.
For the latter, %
multi-modal prompts in ProTrack may wrongly guide the trackers as it changes the data distribution.
For this occasion when the tracking challenges can be solved by RGB trackers, the tracking performance after prompting may be disturbed by auxiliary information, as shown in Fig.~\ref{fig:failed}(b).
Thus, it will be necessary to design more robust prompts in the future, although our current approach can solve a considerable part of the problems.

\section{Conclusion}
In this paper, we propose multi-modal prompts for object tracking, a simple but effective approach to leverage large-scale RGB tracking models for a wide range of downstream multi-modal tracking tasks.
Through applying prompting on multi-modal videos, we adapt the multi-modal tracking tasks to pre-trained RGB trackers.
Thus, by solely modifying the trackers' input, we exploit the most of the discriminative ability from pre-trained RGB trackers to handle the multi-modal tracking challenges.
Promising results on 6 benchmark datasets verify the effectiveness of our proposed ProTrack.

We hope this work will spur further research on multi-modal tracking and provide inspiration to related areas.

\begin{acks}
This work is supported by the National Natural Science Foundation of China under Grant No. 61972188 and 62122035.
\end{acks}


\bibliographystyle{ACM-Reference-Format}
\balance
\bibliography{sample-base}

\end{document}